# WEB CRAWLER STRATEGIES FOR WEP PAGES UNDER ROBOT.TXT RESTRICTION

Piyush Vyas
Department of Information Technology
Sri Vaishnav Institute of Technology and Science
Indore, India.

Akhilesh Chauhan
Department of Information Technology
Sri Vaishnav Institute of Technology and Science
Indore, India.

Tushar Mandge
Department of Information Technology
Sri Vaishnav Institute of Technology and Science
Indore, India
Surbhi Hardikar
Department of Information Technology
Sri Vaishnav Institute of Technology and Science
Indore, India.

*Abstract*

In the present time, all know about World Wide Web and work over the Internet daily. In this paper, we introduce the search engines working for keywords that are entered by users to find something. The search engine uses different search algorithms for convenient results for providing to the net surfer. Net surfers go with the top search results but how did the results of web pages get higher ranks over search engines? how the search engine got that all the web pages in the database? This paper gives the answers to all these kinds of basic questions. Web crawlers working for search engines and robot exclusion protocol rules for web crawlers are also addressed in this research paper. Webmaster uses different restriction facts in robot.txt file to instruct web crawler, some basic formats of robot.txt are also mentioned in this paper.

I.Introduction

Search engines are the soul of the internet which provides all kinds of searches for users. Generally, users enter a keyword for searching for something and got many options in results that are relevant to the search. Search engines provide multiple web pages to surfer for getting the information they want. Web crawlers working for search engines have many versions for different search engines like Googlebot, web spider, indexer, web robot, and so on. Crawler provides indexing to millions of web pages for ranking over search engines. Every webmaster has to specify visiting conditions for the web crawler. That all restrictions, and instructions are mentioned in the robot.txt file. Robot .txt indicates which crawler allows for a web page or which page crawler may not crawl. In this research work, we explain all things in sessions.

## II. Search engine

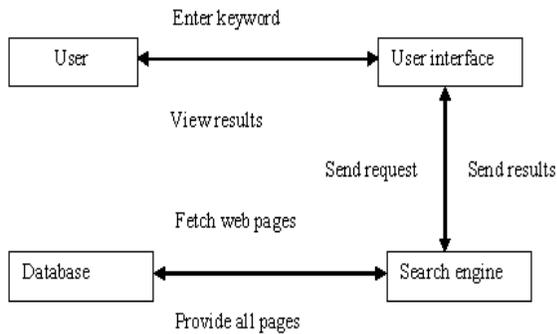

Figure 1 Basic architecture of search engine

A platform that searches different types of domains according to their ranking to make searching convenient is known as a search engine. There are many search engines that provide search facilities for scientific, geographic, commercial, and many other aspects.

A. Working and architecture (See Figure 1):

1. First, net surfer enters a keyword for searching over an interface between them and the search engine.
2. The user interface helps the surfer to interact with the search engine for entered queries.
3. Search engines get the keyword and apply the algorithm to find that from the server database.
4. After getting all information from the database search engine reply to user interface whit results.
5. Finally, a net surfer gets all search results over user interface which they request [1]

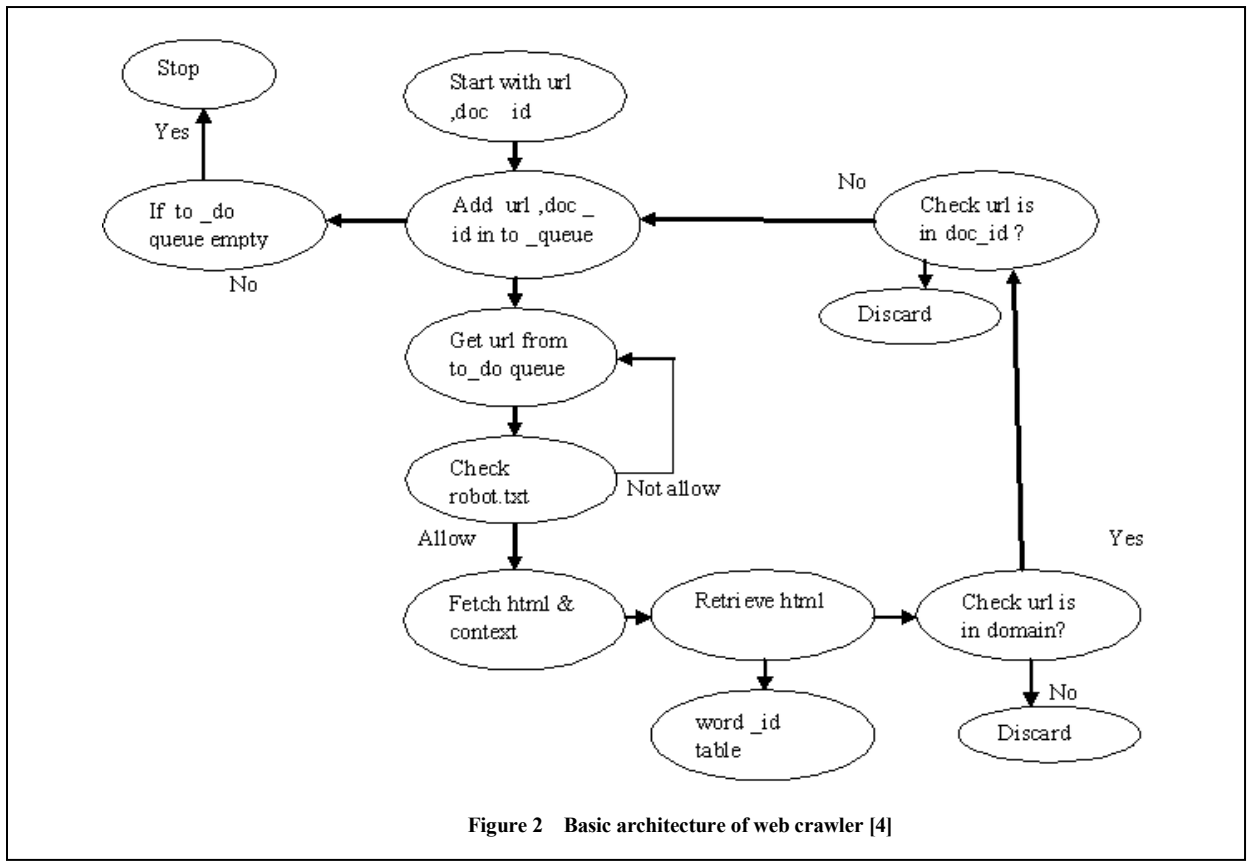

Figure 2  Basic architecture of web crawler [4]

## III. Crawler

Web crawlers originated when the internet came into existence, first web crawlers were used between 1993 to 1996 [2]  A web program that visits websites and automatically downloads pages from the web is known as a web crawler or in other words, a crawler fetches the web pages to index them for search engines to provide a convenient search to surfers [3].

A. Working and architecture (See Figure 2 adopted from [4]):

1. Basically, crawlers analyze URLs and follow their hyperlinks to fetch other URLs of web pages and index all of them. Indexing is the organization of meaningful data (Keywords) that can serve by search engines to retrieve web pages.
2. Crawler gets Url and doc_id to crawl web page. Here doc_id hash table is the storage of the URL of a given domain and provides a separate id to every web document.
3. Then the crawler adds that URL and doc_id to the queue and checks robot.txt for crawler restrictions.
4. If robot.txt allows that crawler to crawl a web page then the crawler fetches the context/HTML and stores that in the doc_id table. If robot.txt does not allow the crawler to crawl that page then the crawler goes back to the queue for further web page's URL.
5. After getting html/context crawler store words in the word_id table and go with html links to crawl them. Word_id table is a collection of keywords and every word has a unique id.
6. If the retrieved HTML URL is in the crawler's current working domain then the crawler checks that status is in the doc_id table if url present in that, then discard that URL, and if not, then add it to the doc_id table. Another thing is if the URL does not belong to the working domain then the crawler discards that immediately.
7. Crawlers continue the above steps till the to_do queue gets empty and after that crawler stops crawling and jumps to another domain [5].

B. Crawler Policies:

a. Revisit policy: crawler works as a function in the web to fetch web pages and it works every second. crawler crawls thousands of pages every second but when a crawler crawls one website, at that time other crawled sites get changes in that context. Changed sites wait for the next crawling and search engines found that the data of the site got outdated. Here freshness and age of the page are two main factors for crawling.

a. Freshness: Newly updated page is fresher than an old crawled page. Suppose P is a web page then the freshness of P at time t is F (P, t) = 1 if P is updated on time t otherwise freshness of P is 0 [6].

ii. Age: age specifies how out of date page P is, age of page P at time t is A (p, t) = 0 if the page update on time t, otherwise (t – modification of P) [6].

b. Uniform policy: In its crawler revisit the collection of all pages at the same frequencies regardless of their rate of changes.

c. Proportional policy: In it, the crawler needs to revisit frequently because pages change more frequently. So here visiting frequency is directly proportional to change.

d. Politeness policy: crawler fetches data in deep and much quicker than human searches, so it affects website performance. To solve this problem robot.txt come into existence. In robot.txt web master declare which part of site crawler can crawl and which one is not. Crawler used a delay parameter in it to extend the crawling time [7].

e. Parallelization policy: A crawler that processes parallel on multiple pages is known as a parallel crawler. The main aim to use it is to maximize the rate of downloading pages and avoid re-downloading of the same web pages [7].

## IV. Robot exclusion protocol

It's a standard in which every host has permission to off-limit crawling. This is done to put a ROBOT.TXT protocol file in the root of the URL hierarchy. When a crawler wishes to fetch any URL, the first crawler read the robot.txt file and the instructions mentioned in it. If the crawler allows by robot.txt file then the crawler fetches the web pages otherwise quit from that session or domain. Robot.txt also mentioned which page has to crawl and which one is not. If there are no robot.txt files then all crawlers are free to fetch every directory of that domain.

A. Format of robot.txt:

Robot.txt has a simple file format and specification: <field>: <optional space><value><optional space>. Here three case incentives for <field> are, User-agent, Allow, disallow and Crawler delay to limit the frequency of crawler visits. User-agent indicates which crawler rule applies to, and allows and Disallow indicates where to crawl and where not. * Sing specify to "all crawler" [8].

Examples:
User-agent: *
Disallow:
It means that all crawlers allow to visit all directories.
User-agent: *
Disallow: /
It means that all crawlers are out of the session.
User-agent: *
Disallow: /images/
It means the crawler does not have permission to fetch the images directory.
User-agent: Googlebot
Disallow: /images/
It means that the Googlebot crawler has not permitted to fetch images directory [8].

## V. Future scope

Lately, the advancement in artificial intelligence (AI) has engaged researchers to automate various tasks utilizing machine learning techniques. Integrating AI in search engines can be revolutionary and recently tech giants like Microsoft have tried to integrate the Chat GPT into the Bing search engine. Moreover, the advanced AI-based techniques, specifically machine earning and deep learning have shown promising results in various domains as e-commerce, social media, fake news, health misinformation, malicious AI bot detections and etcetera. Thus, search engine-based research should start adopting AI techniques to enhance the quality of search-based results. However, many big tech companies are trying to get optimal results but the ethical integration of such automated AI techniques has yet to be discovered.

## VI. Conclusion

The proposed paper shows how the search engine works over the crawler's collected database of visited websites. Our study shows the basic five crawler policies. Crawler works overall policies to revisit the website and to collect keywords for search engines. In this research paper, we studied freshness and age of context and how crawler knows about revisit time for a webpage. Robot.txt restrictions for crawlers directly indicate which area of the site the crawler is permitted to visit and which one is not. All terms and approaches are important to understand for search engine optimization techniques. We put all our effort to understand complex things in an easy way in our paper.